\documentclass[article]{elsarticle}

\usepackage{lineno,hyperref}
\usepackage{framed,multirow}

\usepackage{amssymb}
\usepackage{latexsym}
\usepackage{amsmath}
\usepackage{subfig}
\usepackage{longtable}
\usepackage{afterpage}
\modulolinenumbers[0]

\journal{Digital Signal Processing}

\begin{document}

\begin{frontmatter}

\title{Multispectral image denoising with optimized vector non-local mean filter}
\tnotetext[mytitlenote]{Corresponding authors: }

\author[mymainaddress,mysecondaryaddress]{Ahmed Ben Said \corref{mytitlenote}}
\ead{Tel.:+974 55 748 565; E-mail address: abensaid@qu.edu.qa}
\author[mymainaddress]{Rachid Hadjidj}
\author[kdm]{Kamal Eddine Melkemi}
\author[mymainaddress]{Sebti Foufou}

\address[mymainaddress]{CSE Department, College of Engineering, Qatar University, P.O. Box 2713, Doha-Qatar}
\address[mysecondaryaddress]{LE2i Lab, UMR CNRS 6306, University of Burgundy, BP 47870, 21078, Dijon-France}
\address[kdm]{Department of Computer Science, University of Biskra, 07000, Algeria}

\begin{abstract}
Nowadays, many applications rely on images of high quality to ensure good performance in conducting their tasks. However, noise goes against this objective as it is an unavoidable issue in most applications. Therefore, it is essential to develop   techniques to attenuate  the impact of noise, while maintaining the integrity of relevant information in images. We propose in this work to extend the application of the Non-Local Means filter (NLM) to the vector case and apply it for  denoising multispectral images. The objective is to  benefit from the additional information brought by  multispectral imaging systems. The NLM filter exploits the redundancy of information in an image to remove noise. A restored pixel is a weighted average of all pixels in the image. In our contribution, we propose an optimization framework where we dynamically fine tune the NLM filter parameters and attenuate  its computational complexity by considering only pixels which are most similar to each other in computing a restored pixel. Filter parameters are optimized using Stein's Unbiased Risk Estimator (SURE) rather than using ad hoc means. Experiments have been conducted on multispectral images corrupted with additive white Gaussian noise and PSNR and similarity comparison with other approaches are provided to illustrate the efficiency of our approach in terms of both denoising performance and computation complexity.
\end{abstract}


\end{frontmatter}

\section{Introduction}
\label{intro}
Multispectral imaging systems have emerged as a new technology that is able to deal with various problems encountered with broadband systems.
The development of new cameras and filters enables us to see beyond the visible spectrum such as in the Infrared [700nm-1mm], Ultraviolet [10nm-380nm], X-ray [0.01nm-10nm] spectrums. A multispectral image is a set of several monochrome images of the same scene, each of which is taken at a specific wavelength. Each monochrome image is referred to as a band or channel. A multispectral  image may be seen  as a three dimensional image cube with two spatial dimensions consisting of  the vertical and horizontal axes, and a spectral third dimension. A monochrome image has one spectral band where each  pixel is represented by one scalar value. A multispectral image consists of at least two bands. A pixel is represented by a vector of $P$ components where $P$ is the number of spectral bands. In that sense, a color image may be seen as a multispectral image with three spectral bands. However,  the term "multispectral" is commonly used for images with more than three spectral bands. Images with more than a hundred bands are  commonly called hyperspectral images.
\newline
Using multispectral images is justified by two main reasons. First, narrow spectral bands exhibit more relevant information compared with conventional broadband color and black and white images. Indeed, we obtain a unique spectral signature of the objects being captured. Such information can be used to enhance the accuracy of image processing applications. Second, by using multispectral images, we are able to separate the illumination information from object reflectance, in contrast to broadband images where it is almost impossible to do so. This separated information can now be used to normalize images. For instance, in face recognition applications, near-infrared spectral band can be combined with the visible image. This approach has been widely used to construct more effective biometric systems \citep{raey,1251148,543102}. Thermal infrared images have also been widely used. Thermal infrared sensors detect the heat energy radiated from the face which is independent from the illumination as in the case of  reflectance \citep{raey1,Kong2005103}. Furthermore, thermal infrared is less sensitive to scattering and absorption by smoke or dust and invariant in case of illumination change \citep{Socolinsky200372}. It also allows to reveal anatomical information which is very useful in detecting disguised faces \citep{855246}.
\newline
The quality of a multispectral image has  great implications on the efficiency of image processing applications. Environmental corruption such as noise and blur is a common phenomena for any captured images due to many factors. In particular, a multispectral image can be subject to quality degradation due to the imperfectness of sensors \citep{164271}. Particularly, noise is inevitable in all real broadband and multichannel images. Thus, it is essential to have techniques that ensure noise removal to adequate levels in order to increase performance in many image processing applications such as classification and segmentation \citep{1381,2003IJRS}.\newline
In our work, we propose an improved version of the NLM filter called Optimized Vector NLM (OVNLM), where we take into account the spectral dimension of data.
In fact, the NLM denoising filter exploits the redundancy of information in image. In this paper, we propose an optimization approach to tune the filter parameters. These parameters are optimized using Stein's Unbiased Risk Estimator (SURE) rather than using ad hoc means. Furthermore, we propose a modification to the NLM filter in order to improve its performance. Indeed, the OVNLM is proposed to attenuate the computational complexity of the NLM filter by considering only pixels which are most similar to each other in computing a restored pixel. This attenuation is achieved using a similarity measure based on a probabilistic approach.
\newline
In Section 2, we present an overview of the previous works related to multispectral image denoising and give more details about our contribution. Section 3 summarizes the NLM filter and describes the proposed OVNLM filter. In Section 4, we present experimental results of our framework and  compare them with similar schemes performed on multispectral images. Conclusions are drawn in Section 5.
\section{Related works}
Several techniques have been proposed to tackle the problem of multispectral image denoising.
The work of Luisier et al. \citep{44} represents  the state of the art in multispectral image denoising. Authors proposed a denoising algorithm parameterized as a linear expansion of thresholds \citep{4412968}. Optimization is carried out using Stein's Unbiased Risk Estimator (SURE) \citep{1981,2015,2014}. The thresholding function is point wise and wavelet based. A non-redundant orthonormal wavelet transform is applied on the noisy input image. Next, a vector-valued thresholding of individual multichannel wavelet coefficients is performed. Finally, an inverse wavelet transform is applied to obtain the denoised image. The application of an orthonormal wavelet transform is justified by two main properties. First, assuming a white Gaussian noise in the image domain $\Omega$, its wavelet coefficients remain also Gaussian and are independent between subbands. Second, the Mean Square Error (MSE) in $\Omega$ is equal to the sum of subbands' MSEs.\newline
Another SURE based approach was proposed in \citep{4567}. Authors used a generalized form of shrinkage estimate. The optimal choice of  parameters is based on the minimization of the quadratic risk or MSE that depends on the original data which is unknown. Parameters are chosen so as to minimize the obtained risk. Note that the proposed denoising framework was built around  a wavelet-based approach. Two decomposing schemes were proposed: a decimated M-band wavelet transform and an M-band dual-tree wavelet decomposition. For each case, the associated estimator is obtained.\newline
Another scheme was proposed in \citep{5282540}. The algorithm which jointly removes noise and blur from images is based on the Expectation and Maximization (EM) algorithm \citep{109361}. The noisy signal is decomposed into two independent parts: the first one represents the blurring problem while the second represents the denoising one. The latter is performed in wavelet domain. A Gaussian scale mixture is used to model the probability density of the wavelet coefficients. Besides that, a coregistered auxiliary noise-free image of the same scene is included in the framework to improve the restoration process. In fact, this inclusion provides an extra prior information to the model.
In \citep{6061940}, authors proposed a partial differential equation denoising system  based on the Total Variation (TV) denoising method  used in \citep{tv} which  proposes  to minimize an objective function. For this, authors used the time marching method \citep{Rudin}. The denoising task is then reduced to a partial differential equation (PDE) problem. Authors injected in this PDE problem an auxiliary image as a prior. This approach is justified by the fact that edge directions and texture information of the auxiliary image are similar to those of the noisy image. Thus, a smoothing term that takes into account the contribution of this prior information is added. Although this approach offered better noise smoothing and details conservation, the availability of a reference image as a prior is not straightforward.  \newline
A non-local multidimensional TV model has been recently proposed in \citep{Li2015230}. Authors presented the denoising problem as a minimization of a mean square cost function that depends on a regularization term. The non-local property is not restricted to patches from one band but also to other bands with high correlation. Thus, for a given pixel, the similarity between patches from other bands is considered in the computation of the weight. The multichannel image is first divided into many groups. For a given band, bands with high correlation are grouped together. In addition, the regularization parameters are computed adaptively for each band and derived from the estimated noise standard deviation using the coefficient of the highest frequency wavelet subband. The obtained minimization problem is solved using Bregmanized operator splitting \citep{90746379,sb1,sb2}  which introduces an auxiliary variable. The unconstrained problem is treated  using Bregman iteration method which leads to an update algorithm where Gauss-Seidel and shrinkage methods are used. The proposed framework was jointly used for multichannel image denoising and inpainting. Although the non-local approach offers good denoising performance, it is still computationally expensive and memory space consuming.
\newline
In \citep{6819824}, Zhao et al proposed a denoising framework based on sparse presentation and low rank constraint. Authors analyzed the difference in rank between a clean and a noisy image and concluded that the rank of the clean image is far smaller than the size of the multichannel image. However, this is not true for the noisy image. Thus, an assumption is made: a low rank is a characteristic of a noise free multichannel image. This information is incorporated in the cost function. Furthermore, the cost function requires patch extraction. To avoid the problem of curse of dimensionality and large error, authors suggested to reshape the 3D spectral cube into a 2D matrix by converting each band into a vector, then patches are extracted. The optimization with respect to some variables is carried out by fixing some other variables. The overall complexity of this approach is $O(HLP^2+HLP+L)$ where  $H$ and $L$ are the height and length of the spatial dimension. Although good denoising performance was obtained, this approach doesn't perform well in the presence of high level of noise since it is based on dictionary learning for sparsity representation\newline
Yuan et al. studied in \citep{6558828} the noise in multichannel images, and concluded that there are two types of noise distributions: one distribution in spatial domain and one in spectral domain. Thus, two TV models are used: one applied for multichannel image denoising in spatial domain and the other one is applied in the spectral domain. The two models are both optimized with the split Bregman method where the regularization parameter is selected as the one with highest mean Peak Signal-to-Noise Ratio and Structural Similarity index. Authors studied also the complementary nature of both schemes and concluded that both denoising results can complement each other and that a fusing process can bring additional improvements. By using the metric $Q$ proposed in \citep{5484579}, a fusion scheme between bands from each denoising result is proposed and the final denoised multichannel image is obtained. This approach exhibited good denoising performance but can be improved by adaptively adjusting the regularization parameter on which the denoising performance is highly dependent.
\newline
Yuan et al. in \citep{7054493} proposed also another denoising method where the regularization term in the cost function is often approached by a kernel model. However, this approach has three main drawbacks when applied for multichannel image denoising. First, the spectral information is not considered. Second, since the spatial resolution is lower than the spectral resolution, this approach is inefficient. Finally, noise differs from one band to another. This fact is not considered. Given these challenges, authors suggested two strategies. In the first one, a spectral-spatial kernel model is considered where the spatial and spectral information are simultaneously used. In the second one, noise distributions in spectral bands are considered different and a local kernel is used to balance the contributions between bands. However, the regularization parameter which balances the contribution between the regularization term and fidelity term is not adaptively estimated.
\newline
In \citep{7063939}, authors proposed a denoising framework based on the Bayesian least squares optimization problem. This framework requires the computation of the posterior distribution based on Monte Carlo sampling \citep{montecarlo}. Given the noisy pixel, the procedure starts by choosing some neighbor pixels. Then, the acceptance probability of the sampled pixel given the noisy one is used to decide whether the sampled pixel is to be considered or not. This decision is based on a comparison between the acceptance probability and the random variable drawn from the uniform distribution. After selecting sample pixels, the importance-weighted Monte Carlo posterior estimate is computed using the weighted histogram approach proposed in \citep{wong2011stochastic}, then finally the denoised pixel is obtained.
\newline
Peng et al. proposed in \citep{5414250} a vector version of the bilateral filter. The basic assumption behind this filter is that pixels which have influence on the restored pixels are not just neighbor pixels but neighbor pixels with similar values. Typically, in a similar way to Gaussian filter, bilateral filter is defined as a weighted average of neighbor pixels. However, in order to preserve edges, the difference in value with the neighbor pixels is taken into account. In their work, Peng et al. extended the bilateral filter to the vector form. The dissimilarity measure is now expressed as a multivariate Gaussian function. For simplification purposes and to avoid the computation of the noise covariance matrix, data are projected into subspace using Principle Component Analysis (PCA), and noise variance of individual channels is computed using the median absolute deviation method \citep{Rousseeuw}. However, this ad hoc method makes the results enormously dependent on the choice of filter parameters. \newline
Authors in \citep{6467316,6648714} proposed an optimization framework for the vector bilateral filter using SURE. They  proved that within a neighborhood of a given edge pixel, a high Signal to Error (SER) measure is obtained by maximizing the weight attributed to neighbor pixels with similar values and minimizing the weight given to pixels with significant different values. Authors have also demonstrated that the SER of the vector version of the bilateral filter is always greater than the component wise 2D bilateral filter. The optimization scheme is based on the minimization of the MSE. However, the underlying difficulty of this measure is that it involves the original image which is unknown. MSE is seen as a random variable that depends on the noise. Its expected value is called the risk. To overcome this issue, filter parameters are obtained by minimizing the unbiased risk expression of the SURE estimator. The obtained minimization problem is non-linear and is solved numerically using Sequential Quadratic Programming (SQP). Experiments on color and multispectral images have been conducted and comparison using the Peak Signal to Noise Ratio (PSNR) is presented. \newline
Maggioni et al. \citep{6253256} presented an extension to the BM3D denoising algorithm \citep{4271520} called BM4D. Based on the paradigm of grouping and collaborative filtering, cubes of voxels are stacked and processed in the transform domain which exploits correlation within cubes and the non-local correlation between the corresponding voxels of different cubes. This approach leads to an effective separation between signal and noise through coefficient shrinkage. \newline
Peng et al. \citep{6909773} proposed the TDL algorithm. Authors focused on the spatial non-local similarity and the spectral correlation of multispectral images. A non-local tensor dictionary learning model is developed. This model is constrained by group-block sparsity. In addition, the proposed model is decomposed into a series of low-rank tensor approximation problems. These problems are approached using higher-order statistics.
\newline
Manjon et al. \citep{466} have recently proposed a new algorithm for multispectral image denoising based on the Non-Local Mean (NLM) filter \citep{46}. NLM filter is designed so that it takes advantage of the redundancy exhibited in the image. This redundancy is no longer pixel based but window based. In other words, every small window centered on a pixel is supposed to have many similar windows. These windows can be located anywhere in the image domain $\Omega$ and are no longer restricted to the neighborhood. In the multispectral framework, information from various bands are combined and a new weight is proposed. Also NLM filter is very effective, it is highly dependent on the choice of three parameters: the radius of the search window, the radius of the neighborhood window and a smoothing parameter that controls the degree of the smoothing. The latter is very important. Indeed, with a small value, little noise will be removed. On the other hand, with a high value, the image will be blurred. Authors have set these parameters manually.
\newline
Motivated by the successful applications of NLM filter in image denoising and details preservation compared to other filters (e.g. bilateral), we propose a modified version of the NLM filter called Optimized Vector NLM (OVNLM) filter. Our contribution consists in: 
\begin{itemize}
\item Proposing a vector version of the NLM filter where we take into consideration the additional information brought by the spectral imaging system. 
\item Automatic tuning of the filter parameters. Unlike the ad hoc method proposed in \citep{466}, we use an optimization approach to properly choose these parameters in a way that guarantees better denoising performance. 
\item Reducing the computation complexity. The main advantage of NLM filter is its non-local property which means each pixel is influenced by all pixels in the image. This comes unfortunately with more computation burden. We alleviate this burden by proposing a similarity measure used to decide whether we should take the pixel contribution or not during the pixel restoration process. Our experimental results demonstrate that this approach not only reduces the computation time but also ensures good denoising results. 	
\end{itemize}
We prove through quantitative evaluation the advantages of the proposed method compared to other denoising algorithms derived from the classic NLM filter as well as from other theories. Indeed, our method achieves better denoising performance compared to other algorithms. Furthermore, we show how OVNLM is capable to preserve image details while conserving its non-local property and ensuring acceptable computational efficiency.   
\section{Optimized vector Non-Local Mean filter for multispectral image denoising: OVNLM}
We consider the following additive noise model:
\begin{equation}
I_{in}(s)=I_{or}(s)+N
\label{model_eq}
\end{equation}
Where $I_{in}(s)$ and $I_{or}(s)$ are the noisy and original pixels respectively, $N$ is the Gaussian noise and  $s \in \Omega$ is the pixel coordinates in the spatial domain.

\subsection{Non-Local Mean filter}
The basic assumption behind the definition of the NLM filter is that we need to take advantage of the high degree of redundancy in the image: the neighborhood of a pixel $s$ is any set of pixels $p$ in the image domain $\Omega$ such that a local window surrounding $s$ is similar to the local window surrounding $p$ \citep{46}.
The general case of NLM filter is given by:
\begin{equation}
I_{out}(s)=\sum_{p\in \Omega} \omega(s,p)I_{in}(p)
\end{equation}
$\omega(s,p)$  is the weight calculated for each pixel. It is computed based on a similarity measure between pixels in position $s$ and $p$. $\omega(s,p)$ satisfies the following constraints:
\begin{equation}
\begin{tabular}{ l }
\(0\leq \omega(s,p) \leq 1\)\\
\( \sum_{p\in \Omega}\omega(s,p)=1\)
\end{tabular}
\end{equation}
The similarity between two pixels $s$ and $p$ is measured as a decreasing function of the Gaussian weighted Euclidean distance $\left\|\cdot\right\|^{2}_{2,a}$, where $a>0$ is the standard deviation of the Gaussian kernel. Let $N(s)$ and $N(p)$ be the pixel vectors of the gray level intensity within a squared neighborhood centered at positions $s$ and $p$ respectively.
\begin{equation}
\omega(s,p)=\frac{1}{C_i}exp\left( -\frac{\left\|N(s) - N(p) \right\|^2_{2,a}}{h^2}\right)
\end{equation}
$h^2$ acts as a smoothing parameter. $C_i$ is a normalization constant which ensures that $\sum_{p\in \Omega}\omega(s,p)=1$.
\begin{equation}
C_i=\sum_{p \in \Omega}exp\left( -\frac{\left\|N(s) - N(p) \right\|^2_{2,a}}{h^2} \right)
\end{equation}
The Gaussian weighted Euclidean distance is given by:
\begin{equation}
\left\|N(s) - N(p) \right\|^2_{2,a}=\sum_{k\in K}G_{a}(k)(N(s-k)-N(p-k))^{2}
\end{equation}
where $K$ is a local window and $G_{\alpha}(k)$ is defined as:
\begin{equation}
G_{\alpha}(k)=\frac{1}{2\pi a^2}exp\left( -\frac{k_1^2+k_2^2}{2a^2} \right), k=(k_1,k_2)
\end{equation}
Thus, we can distinguish two main characteristics: the restored pixel is obtained by taking into account the contribution of pixels in the whole image and the weight computation is based on the similarity between local windows. Such characteristics have triggered researchers to design various novel methods \citep{46}.

\subsection{Vector NLM filter}
To take advantage of the additional information brought by the spectral dimension, we extend the NLM filter to the vector case.
In the multispectral context, we have the reflectance intensity at a given position in different spectral bands. Thus, we are operating on a set of pixel vectors
$I=\{(I_s) /{s \in \Omega}\}$. We define the vector NLM (VNLM) filter as:
\begin{equation}
I_{out}(s)=\sum_{p\in \Omega} \omega(s,p)I_{in}(p)
\label{eq_vonlm}
\end{equation}
where  the new formulation of the weight between two pixels at position $s$ and $p$ is defined as:
\begin{equation}
\begin{tabular}{l}
\(\omega(s,p)=\) \( \frac{1}{C_i}exp \biggl(\frac{-1}{h^2}\sum_{k\in K}(I_{in}(s-k)-I_{in}(p-k))^{T}\Phi^{-1}\)\\
\(\cdot (I_{in}(s-k)-I_{in}(p-k)) \biggr)  \)
\end{tabular}
\end{equation}
If $\Phi=Id$, where $Id$ is the identity matrix, we get the classical Euclidean distance.
\begin{equation}
\begin{tabular}{l}
\(C_i= \sum\limits_{p \in \Omega}exp\biggl(-\frac{1}{h^2}\sum_{k\in K}(I_{in}(s-k)-I_{in}(p-k))^{T}\Phi^{-1}\)\\
\( \cdot (I_{in}(s-k)-I_{in}(p-k))\biggr) \) \\
\end{tabular}
\end{equation}


\subsection{Optimization framework for vector NLM}
In our framework design, we target two main objectives: optimize the parameters of the filter and reduce its computation complexity. First we use both the classical Euclidean distance $\left\|\cdot\right\|_2^2$ and Mahalanobis distance $\left\|\cdot\right\|^2_{\Phi}$ where $\Phi$ is a covariance matrix. In addition, we preselect for each pixel a subset of the most similar pixels based on a probabilistic similarity measure. \newline
The filter depends on two parameters: the smoothing parameter $h$ and the covariance matrix $\Phi$. Thus, we have:
\begin{equation}
\begin{tabular}{ l l l}
\(I_{out}(s)=f(I_{in}(s),\Theta)\)&\(with\)&\(\Theta=(h,\Phi)\)
\end{tabular}
\end{equation}
where $f$ is a nonlinear estimator and $\Theta$ is the filter parameter.\newline
Our aim is to optimize $\Theta$ so that we can ensure the choice of the optimal parameters in order to obtain the best denoising result. The performance of the estimator is generally evaluated using the mean square error (MSE):
\begin{equation}
MSE=\frac{1}{HL}\sum_{s\in \Omega}\left\| I_{out}(s) - I_{or}(s)\right\|^2
\label{eq_mse}
\end{equation}
However, the problem of such estimator is that the ground truth image $I_{or}(s)
$ is unknown. MSE can be seen as a random variable of the noise. Its expected value is designated as the Risk $R_\theta$ and expressed as:
\begin{equation}
R_\theta=E(MSE)
\end{equation}
The problem of estimating the risk without the need to have the underlying image $I_{or}(s)$ is approached by Stein's Unbiased Risk Estimator (SURE) \citep{4567,44}. Thus, we have \citep{1981}:
\begin{equation}
\begin{tabular}{l}
\(E\left( \left\|I_{out}(s) - I_{or}(s)\right\|^2\right)= \)\\
\(E\left( \left\|I_{out}(s)\right\|^2 \right)-2E\left(I_{out}(s)^T I_{or}(s) \right)+E\left( \left\|I_{or}(s)\right\|^2 \right) \)
\label{eq_nlm1}
\end{tabular}
\end{equation}
and:
\begin{equation}
\begin{tabular}{ l l l }
\(E\left(I_{out}(s)^T I_{or}(s)\right) =E\left(f(I_{in}(s),\Theta)^T (I_{in}(s)-n_s)\right)\)\\
\\
\(  \)\(= E\left(I_{out}(s)^T I_{in}(s)\right) - E\left(f(I_{in}(s),\Theta)^T n_s \right)   \) \\
\end{tabular}
\label{eq_nlm2}
\end{equation}
$T$ is the transpose operator. If we consider zero mean multivariate Gaussian noise, we get \citep{44}:
\begin{equation}
E\left( f(I_{in}(s),\Theta)^T n_s\right)=E\left(trace\left\{\Psi^T\bigtriangledown_{I_{in}(s)}f(I_{in}(s),\Theta) \right\}\right)
\end{equation}
where $\Psi$ is the noise covariance matrix.\\
By combining eq. \ref{eq_nlm1} and eq. \ref{eq_nlm2}, we end up with an expression without $I_{or}(s)$:
\begin{equation}
\begin{tabular}{l}
\(E\left( \left\|I_{out}(s) - I_{or}(s)\right\|^2\right)=E\left(\left\|I_{out}(s)-I_{in}(s)\right\|^2\right) \)\\
\\
\(- trace(\Psi)+2E\left(trace\left\{\Psi^T\bigtriangledown_{I_{in}(s)}f(I_{in}(s),\Theta) \right\}\right)\)
\end{tabular}
\end{equation}
Therefore, the risk $\hat{R_\theta}$ is the unbiased risk estimator of MSE in eq. \ref{eq_mse} and is given by:
\begin{equation}
\begin{tabular}{l}
\(\hat{R_\theta}=\frac{1}{HL}\sum\limits_{s\in \Omega}E\left(\left\|I_{out}(s)-I_{in}(s)\right\|^2\right) - trace(\Psi)\)\\
\\
\(+2\frac{1}{HL}\sum\limits_{s\in \Omega}E\left(trace\left\{\Psi^T\bigtriangledown_{I_{in}(s)}f(I_{in}(s)) \right\}\right)\)
\end{tabular}
\label{eq_r}
\end{equation}
where $\bigtriangledown_{I_{in}(s)}f(I_{in}(s))=J_{f(I_{in}(s))}$ is the Jacobian matrix with respect to $I_{in}(s)$.
$J_{f(I_{in}(s))}$ is given by \citep{6648714}:
\begin{equation}
\begin{tabular}{l}
\((J_{f(I_{in}(s))})_{i,j}=\frac{\partial f_{i}(I_{in}(s),\theta)}{\partial I_{in}(s_j)}= \)\\
\(\frac{\sum\limits_{p \in \Omega}\frac{\partial \chi(p)}{\partial I_{in}(s_j)}I_{in}(s_i)+\delta_{i,j}}{\sum\limits_{p\in \Omega}\chi(p)}- \frac{\left(\sum\limits_{p \in \Omega} \frac{\partial \chi(p)}{\partial I_{in}(s_j)}\right) \left( \sum\limits_{p\in \Omega}\chi(p)I_{in}(s_j)\right) }{ \left( \sum\limits_{p\in \Omega }\chi(p)\right)^2 } \)\\

\end{tabular}
\end{equation}
where $\delta_{i,j}$ is the delta function and $\chi(p)$ is defined as:
\begin{equation}
\begin{tabular}{l}
\(\chi(p)=exp\biggl(-\frac{1}{h^2}\sum_{k\in K} (I_{in}(s-k)-I_{in}(p-k))^{T}\Phi^{-1}\)\\
\( \cdot  (I_{in}(s-k)-I_{in}(p-k))\biggr)\)
\end{tabular}
\end{equation}
With the derivation of $\chi(p)$ (see appendix), 
we formulate the problem of vector NLM filter as a constrained optimization problem:
\begin{equation}
\left\{
\begin{tabular}{l}
\((h_{opt},\Phi_{opt})=\arg\min_{h,\Phi}(\hat{R}(h,\Phi))\)\\
\\
\(s.t.:\)\( \) \( h>0\)\( , \) \(\Phi\geq 0\)
\end{tabular}
\right.
\end{equation}
Note that in case of using the Euclidean distance, the only parameter to be optimized is $h$.

\subsection{Relevant pixel selection}
\begin{figure}[!b]
	\centering
	\includegraphics[scale=.55]{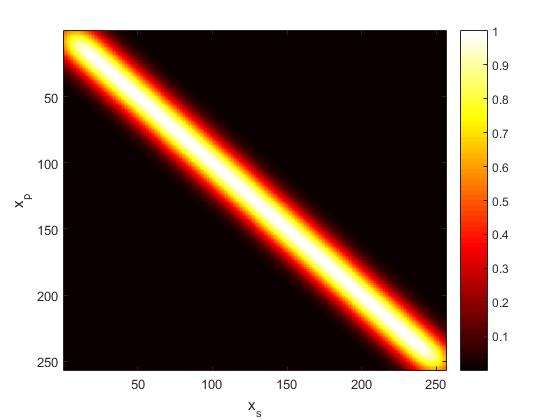}
	\caption{Visualization of the intensity similarity measure with $\sigma^2$=100. Brighter values indicate higher similarity. Values are mapped to the range [0,1] for visualization purpose }
	\label{heat}
\end{figure}

If we go back to eq. \ref{eq_vonlm}, we can clearly see that in order to restore every pixel, we need to go through every other pixel in the domain $\Omega$. This is obviously a very time consuming process.
To attenuate the computation complexity of the proposed VNLM filter, we propose to preselect for each processed pixel, a set of most relevant pixels based on the  similarity measure proposed in \citep{427}.  This measure is based on a probabilistic approach to compute the similarity between two pixels based on the noise distribution. \newline In the grayscale case, the similarity measure is defined as:
\begin{equation}
\begin{tabular}{l}
\(S(x_s,x_p)=\frac{1}{4\sigma|\Omega|\sqrt{\pi}}\cdot \)\\
\(exp(-\frac{(x_s-x_p)^2}{4\sigma^2})\biggl(erf(\frac{2x_{s0}-x_s-x_p}{2\sigma})+erf(\frac{x_s+x_p}{2\sigma})\biggr)\)
\end{tabular}
\label{eq_sim}
\end{equation}
where $x_{s0}$ is the maximum value of the true intensity, $|\Omega|$ is a constant, and $erf(\cdot)$ is the error function defined as:
\begin{equation}
erf(x)=\frac{2}{\sqrt{\pi}}\int^x_0 e^{-t^2}dt.
\end{equation}
Fig. \ref{heat} illustrates the form of the similarity measure for $\sigma^2$=100.
For a given $x_s$, eq. \ref{eq_sim} illustrates a Gaussian function. We consider that all values beyond the width of 1/$\varsigma$ of maximum ($x_p \pm 2\sqrt{2ln(\varsigma)}\sigma$) are zeros.
\noindent In the case of RGB color images, the similarity between two pixels $I_s=[r_s,g_s,b_s]$ and $I_p=[r_p,g_p,b_p]$
is defined as:
\begin{equation}
S(I_s,I_p)=S(r_s,r_p)\cdot S(g_s,g_p)\cdot S(b_s,b_p)
\end{equation}
We generalize this similarity measure for the multispectral case, such that the similarity measure between $I(s)=(I(s_i)_{i=1\ldots P})$ and $I(p)=(I(p_i)_{i=1\ldots P})$ is defined as:
\begin{equation}
S(I_s,I_p)=\prod^{P}_{i=1}S(I(s_i),I(p_i))
\end{equation} 
The proposed Optimized  VNLM (OVNLM) filter becomes:
\begin{equation}
I_{out}(s)=\sum_{\substack{p\in \Omega\\S(I_{in}(s),I_{in}(p))\neq 0}} \omega(s,p)I_{in}(p)
\label{eq_vnlm}
\end{equation}

\subsection{OVLNM algorithm}
The proposed approach is detailed in what follows.
We solve the constrained non-linear optimization problem using Sequential Quadratic Programming. Given a noisy image, the noise covariance matrix is estimated with the standard median absolute deviation method \citep{Huber}. The diagonal elements of $\Psi$ are calculated as follows:
\begin{equation}
\Psi(i,i)=\biggl(1.4826 \times median(|I_{in,i}-median(I_{in,i})|)   \biggr)^2
\end{equation}
$i=1...P$. The off-diagonal elements are defined as:
\begin{equation}
\begin{tabular}{l}
\(\Psi(i,j)= \frac{1.4826^2}{4ab}\biggl[median\biggl(|(aI_{in,i}+bI_{in,j})- median(aI_{in,i}+bI_{in,j})|\biggr)^2 \)\\
\(- median\biggl(|(aI_{in,i}-bI_{in,j})- median(aI_{in,i}-bI_{in,j})|\biggr)^2 \biggr]\)
\end{tabular}
\end{equation}
where:
$a=(\Psi(i,i))^{-1/2}$ and $b=(\Psi(j,j))^{-1/2}$ and $i,j=1...P$.
We minimize the risk value based on an optimal choice of parameters $\Theta=(h,\Phi)$ until we reach the maximum number of iteration $iter\_max$ or the risk value decreases below a preset threshold $\xi$. We implemented this approach in Matlab (R2015a). The minimization is conducted using the function $fmincon$ with the risk as an objective function to minimize and SQP as the optimization approach.
We use a neighborhood window of $7 \times 7$ and we set up $\varsigma=100$.
\begin{table}[h!]
	\centering
	\begin{tabular}{l}
		\hline\noalign{\smallskip}
		\textbf{Input}($I_{in}(s)_{s \in \Omega}), \Psi$\\
		\textbf{Output} Optimal $(I_{out}(s)_{s \in \Omega})$ with minimal $\hat{R}_{h,\Phi}$\\
		\noalign{\smallskip}\hline
		\textbf{1}- Initialize $\Psi$, $h$, $\Phi$, iter=0, maximum iteration number \\
		\( \) $iter\_max$  and stopping threshold $\xi$\\
		\textbf{2}- Iteration: \textbf{do}\\
		\(     \)     a-  Calculate $(I_{out}(s)_{s \in \Omega})$  using Eq. \eqref{eq_vnlm} and \\
		\(     \)     b-  Calculate $R_{iter}$ using \eqref{eq_r}\\
		\(     \)     c-  $iter=iter+1$\\
		\(     \)     d-  Update $h$ with SQP\\
		\(     \)     e-  Update $\Phi$ with SQP\\
		\(     \)     f-  Compute $R_{iter+1}={\hat{R}}(h_{iter+1},\Phi_{iter+1}$)\\
		\textbf{While} ($iter\prec iter\_max$ or $R_{iter}-R_{iter+1}\preceq \xi$)\\
		\noalign{\smallskip}\hline
	\end{tabular}
	\label{nlm_algo}
\end{table}


\section{Experiments}
\subsection{Data sets}
\begin{figure}[!h]
	\centering
	\includegraphics[scale=.35]{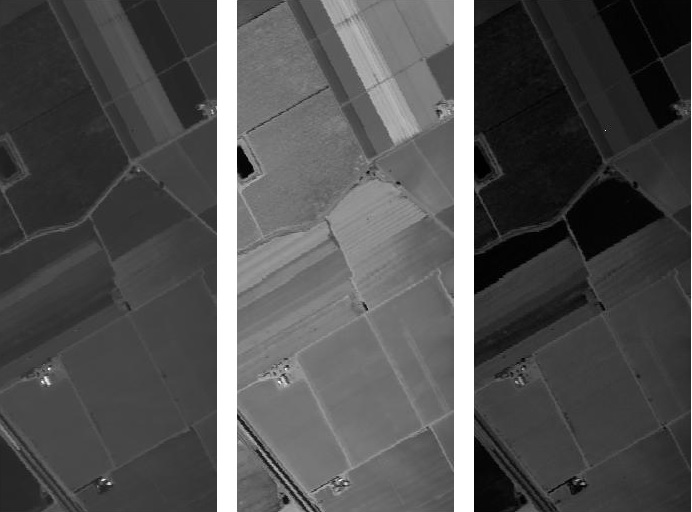}
	\caption{Multispectral Salinas Valley images}
	\label{salinas}
\end{figure}

\begin{figure}[!h]
	\centering
	\includegraphics[scale=.32]{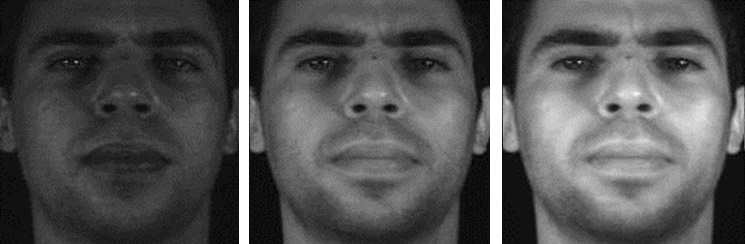}
	\caption{Multispectral images for $Subject_{1}$ in 480nm, 560nm and 720nm}
	\label{ms2}
\end{figure}

\begin{figure}[!h]
	\centering
	\includegraphics[scale=.30]{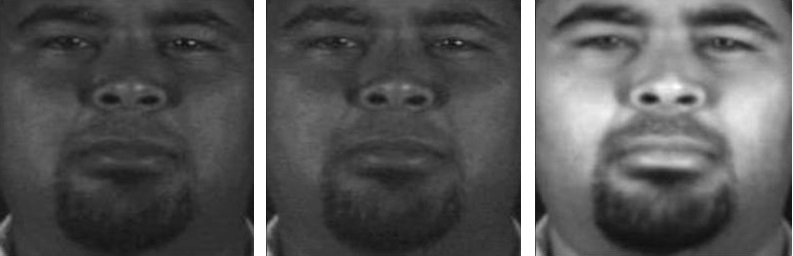}
	\caption{Multispectral images for $Subject_{2}$  in 480nm, 560nm and 720nm}
	\label{ms}
\end{figure}

To assess the performance of our approach, we conducted experiments on real world multispectral images. In one of the experiments, we used the Salinas scene collected using the Airborne Visible Infra-Red Imaging Spectrometer (AVIRIS) at NASA's Jet Propulsion Laboratory\footnote{http://aviris.jpl.nasa.gov/}.  Sample bands are shown in Fig. \ref{salinas}. The Salinas Valley image is  a high spatial resolution image consisting of a  collection of 224 spectral band images taken over Salinas Valley California  in the range from 0.4$ \mu$m to 2.5$ \mu$m at a resolution of 3.7 meters per pixel.  Spectral Bands [108-112], [154-167] and 220 are discarded due to water absorption and noise. Before processing, images are resized  to  $216 \times 216$ pixels.  In another experiment we used  multispectral face images from  the IRIS Lab database  at the University of Tennessee \cite{1640494}.  The IRIS Lab database was built between August 2005 and March 2006 and consists of 2624 multispectral face images taken along the visible spectrum in addition to thermal images with a resolution of $640\times 480$ pixels. RGB images are also generated with a resolution of $2272 \times 1704$ pixels. These images are taken in different lighting conditions: Halogen light, daylight and fluorescent light. The total size of the database is 8.91GB. A total of 82 participants were involved from different genders (76\% male, 24\% female), ethnicities as depicted in Table \ref{tab_eth}, ages, facial expression, genders and hair characteristics. We conducted experiments on 8 multispectral images referred to as $Subject_{i}$  $(i=1 \ldots 8)$. Figures \ref{ms2} and \ref{ms} illustrate the multispectral image samples taken with Halogen light  and used in our experiments.
\renewcommand{\arraystretch}{1}
\begin{table}[!h]
	\caption{Ethnicity percentage in IRIS $M^3$ database}
	\begin{center}
		\begin{tabular}{ c  c  c  c  c }
			\hline
			& Caucasian & Asian & Asian Indian & African descent \\
			\hline
			\% & 57\% & 23\% & 12\% & 8\% \\
			\hline
		\end{tabular}
	\end{center}
	\label{tab_eth}
\end{table}
\subsection{Evaluation results}
\begin{figure}[!h]
	\centering
	\includegraphics[scale=.45]{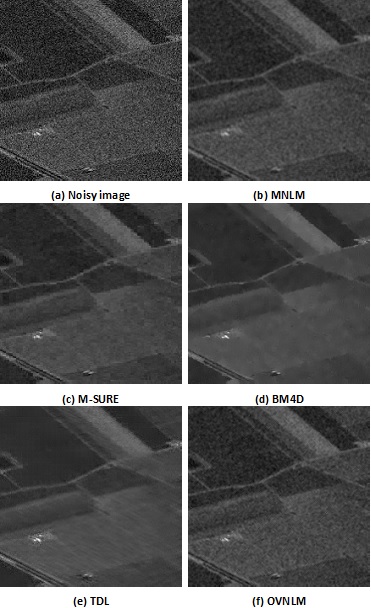}
	\caption{Salinas Valley denoising results: input PSNR=19 dB: (a) Noisy image (b) MNLM \citep{466}: PSNR=30.7 dB (c) The multichannel SURE-LET (M-SURE) \citep{44}: PSNR=29.8 dB, (d) BM4D \citep{6253256}: PSNR=28.7 dB (e) TDL \citep{6909773}: PSNR=25.9 dB (f) OVNLM: PSNR=31.8 dB}
	\label{salina_results}
\end{figure}

\begin{figure}[!h]
	\centering
	\includegraphics[scale=.5]{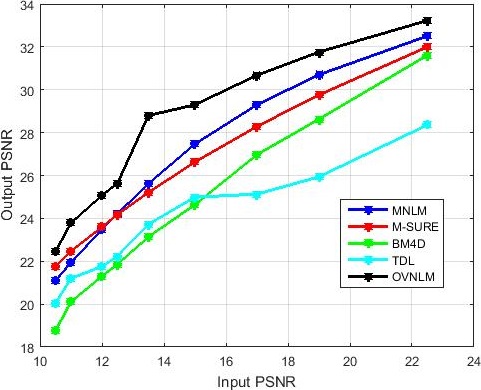}
	\caption{Performance of five denoising algorithms on Salinas multispectral image captured by AVIRIS sensor: MNLM \citep{466}, the multichannel SURE-LET (M-SURE) \citep{44}, BM4D \citep{6253256}, TDL \citep{6909773} and the proposed method OVNLM. Output results are evaluated at different noise levels. OVNLM presents the best performance}
	\label{salina_curves}
\end{figure}

\begin{figure}[!h]
	\centering
	\includegraphics[scale=.45]{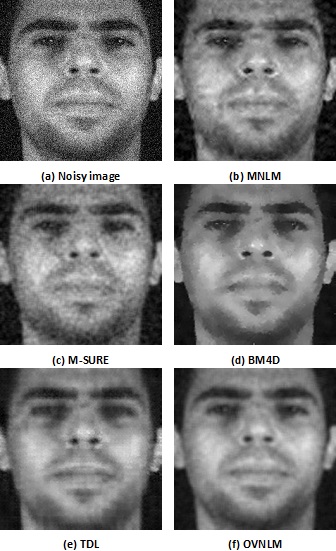}
	\caption{$Subject_{1}$ denoising results with input PSNR=19 dB: (a) Noisy image (b) MNLM \citep{466}: PSNR=27.8 dB (c) The multichannel SURE-LET (M-SURE) \citep{44}: PSNR=31.1 dB, (d) BM4D \citep{6253256}: PSNR=27.4 dB (e) TDL \citep{6909773}: PSNR=27.1 dB (f) OVNLM: PSNR=32.7 dB}
	\label{subject1}
\end{figure}

\begin{figure}[!h]
	\centering
	\includegraphics[scale=.45]{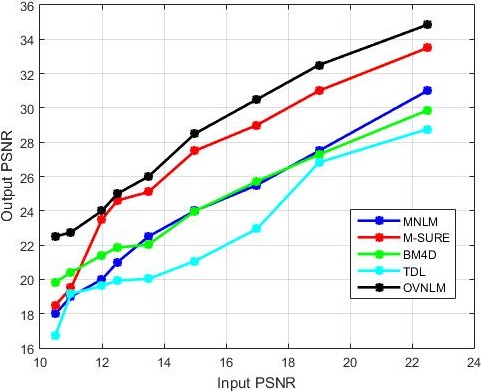}
	\caption{Denoising performance of five denoising algorithms applied on $Subject_{1}$ multispectral image from by IRIS database: MNLM \citep{466}, the multichannel SURE-LET (M-SURE) \citep{44}, BM4D \citep{6253256}, TDL \citep{6909773} and the proposed method OVNLM. Output results are evaluated at different noise levels. OVNLM presents good denoising performance with distinctive results in presence of high noise levels}
	\label{curve_subject1}
\end{figure}

\begin{figure}[!h]
	\centering
	\includegraphics[scale=.5]{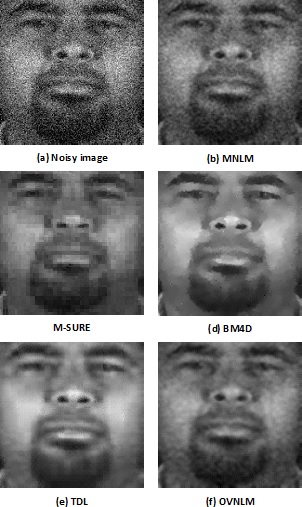}
	\caption{$Subject_{1}$ denoising results with input PSNR=19 dB: (a) Noisy image (b) MNLM \citep{466}: PSNR=27.7 dB (c) The multichannel SURE-LET (M-SURE) \citep{44}: PSNR=28.4 dB, (d) BM4D \citep{6253256}: PSNR=27.5 dB (e) TDL \citep{6909773}: PSNR=25 dB (f) OVNLM: 33.07 dB}
	\label{subject2}
\end{figure}

\begin{figure}[!h]
	\centering
	\includegraphics[scale=.5]{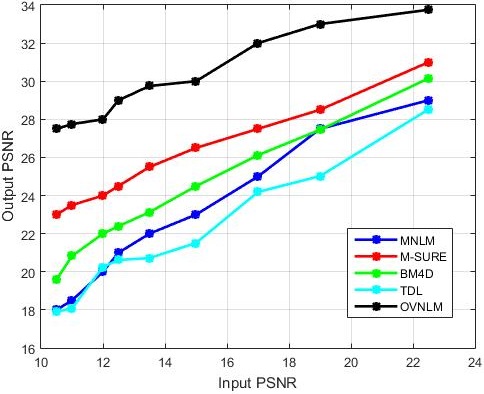}
	\caption{Denoising performance of five denoising algorithms applied on $Subject_{2}$ multispectral image from by IRIS database: MNLM \citep{466}, the multichannel SURE-LET (M-SURE) \citep{44}, BM4D \citep{6253256}, TDL \citep{6909773} and the proposed method OVNLM. Output results are evaluated at different noise levels. OVNLM presents good denoising performance with distinctive result  in presence of high noise levels}
	\label{curve_subject2}
\end{figure}

\begin{figure}[!h]
	\centering
	\includegraphics[scale=.34]{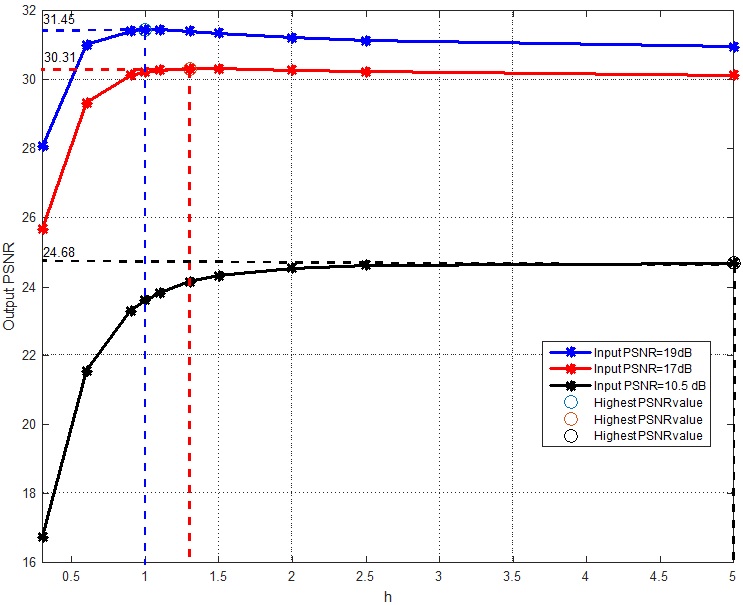}
	\caption{PSNR variation with respect to $h$ with different level of input PSNR: variation of the output PSNR demonstrate the dependance of the filter on parameter h implying also  dependence on parameter $\Phi$ which confirms the importance of applying an optimization approach to tune these parameters. }
	\label{curve_h}
\end{figure}

\begin{figure*}[!tbp]
		\centering
		\subfloat[$Subject_{3}$ ]{\includegraphics[width=0.45\textwidth]{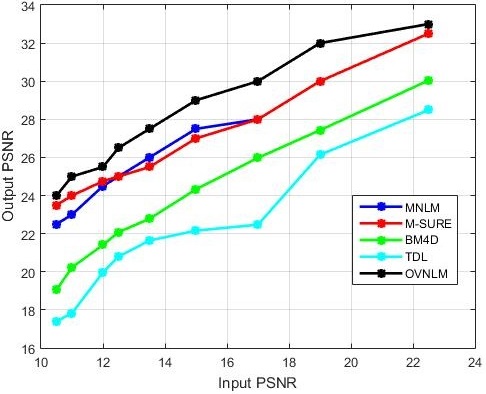}\label{subject3}}
		\hfill
		\subfloat[$Subject_{4}$ ]{\includegraphics[width=0.45\textwidth]{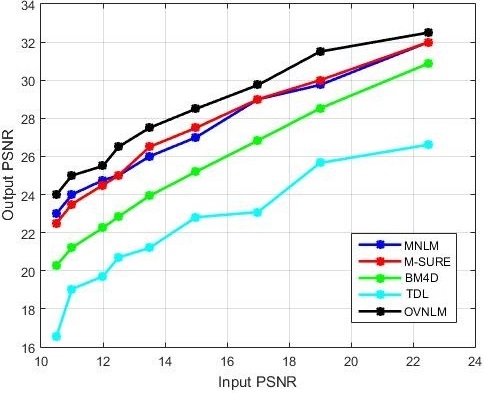}\label{subject4}}
		\quad
		\subfloat[$Subject_{5}$ ]{\includegraphics[width=0.45\textwidth]{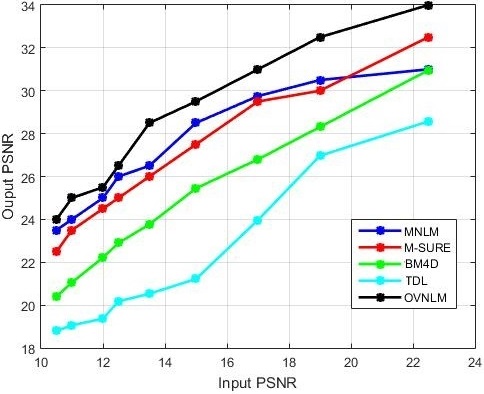}\label{subject5}}
		\hfill
		\subfloat[$Subject_{6}$ ]{\includegraphics[width=0.45\textwidth]{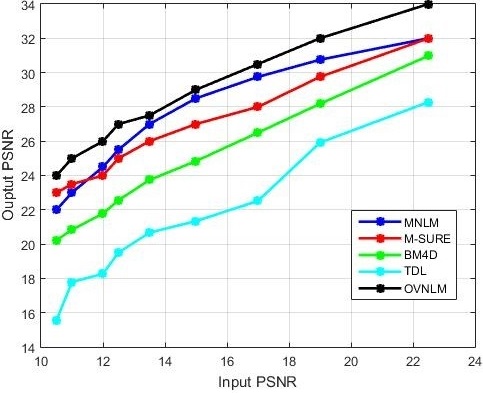}\label{subject6}}
		\quad
		\subfloat[$Subject_{7}$ ]{\includegraphics[width=0.45\textwidth]{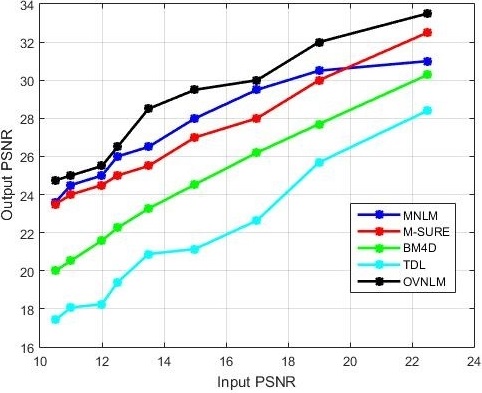}\label{subject7}}
		\hfill
		\subfloat[$Subject_{8}$ ]{\includegraphics[width=0.45\textwidth]{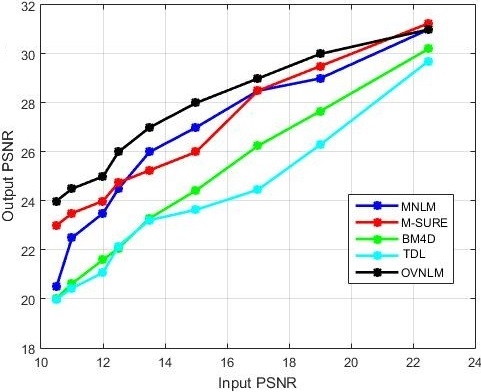}\label{subject8}}
		\caption{PSNR variations of five algorithms applied for denoising $Subject_{i}$  $(i=3 \ldots 8)$ multispectral face images from IRIS Lab database. OVNLM presents the best performance compared to MNLM \citep{466}, the multichannel SURE-LET (M-SURE) \citep{44}, BM4D \citep{6253256}, TDL \citep{6909773}. Some algorithms fail to present good performance such as TDL ($Subject_4$, $Subject_5$) with high level of noise, OVNLM ensures stable performance with all noise levels. }
		\label{sub1_8}
	\end{figure*}

We applied the proposed OVLNM approach and compared its denoising performance with several state of the art multispectral image denoising algorithms: MNLM \citep{466}, the multichannel SURE-LET (M-SURE) \citep{44}, BM4D \citep{6253256} and TDL \citep{6909773}.
Note that MNLM is also inspired from the NLM filter and adapted for multispectral image denoising with choice of parameters conducted using ad hoc means.
Experiments on multispectral image denoising are conducted by contaminating original images with an additive Gaussian noise at different levels, then  denoising algorithms are applied on the noisy images. We use the Peak Signal to Noise ratio (PSNR) expressed in dB and the Structure Similarity Index Measure (SSIM):	
\begin{equation}
PSNR=10log_{10}\left[ \frac{\underset{i,j}{max}\left\{I_{or}(i,j) \right\} ^{2}  }{\frac{1}{MN} \sum_{i=1}^{M}\sum_{j=1}^{N} \left| I_{or}(i,j)- I_{out}(i,j)  \right|^{2}         }\right]
\end{equation}
\begin{equation}
SSIM(x,y)=\frac{(2\mu_x\mu_y+c_1)(2cov(x,y)+c_2)}{(\mu_x^2+\mu_y^2+c_1)(\sigma_x^2+\sigma_y^2+c_2)}
\end{equation}
where $I_{or}(i,j)$ and $I_{out}(i,j)$ are pixel values at position $(i,j)$ in the original and output images respectively.
PSNR is the ratio of the maximum possible value of the signal in term of its power and the power of the distortion caused by the noise. PSNR is expressed in the logarithmic decibel scale. The higher the PSNR is, the better is the result. SSIM is an index that measures the similarity between two images $x$ and $y$ with $\mu_x$, $\mu_y$, $\sigma_x$ and $\sigma_y$ are respectively the mean of image $x$, the mean of image $y$, the standard deviation of image $x$ and the standard deviation of image $y$. $c_1$ and $c_2$ are two stabilization constants and $cov(x,y)$ is the covariance of $x$ and $y$. The best result is highlighted by the highest value of SSIM.
\newline
We show in Fig. \ref{salina_results} the denoising results of the Salinas image with PSNR=19 dB. We notice that over-smoothness is witnessed with M-SURE, BM4D and more with TDL. Some patterns in the images are better preserved particularly with MNLM and OVNLM. Indeed, one of the intrinsic property of an NLM based method is its ability to preserve image details beside its good performance even with high noise level. These characteristics are now strengthened with the fine tuning of the parameters.
In our experiments, we noticed that more detail preservation can be obtained with TDL by reducing the voxel block size but this comes with a loss in term of PSNR.  Fig. \ref{salina_curves} presents the PSNR of denoising results at different level of input PSNR. We can clearly see that OVNLM exhibits the best performance with high PSNR (low noise level) and also with low PSNR (high noise level). This performance is confirmed with the SSIM results presented in Tab. \ref{tab_salinas}.  
\begin{table}[!h]
	\caption{SSIM results for Salinsas Valley multispectral image. The best results are bold. OVNLM exhibits good performance and outperform other methods in presence of high noise level}
	\begin{center}
		\begin{tabular}{c c c c c c}
			\hline
			Input PSNR & MNLM & M-SURE & BM4D & TDL & OVNLM \\
			\hline
			22.5 & 0.82 & 0.87 & \textbf{0.88} & 0.85 & 0.83 \\
			19   & 0.77 & 0.80 & \textbf{0.83} & 0.80 & 0.77 \\
			17   & 0.75 & 0.74 & \textbf{0.78} & 0.77 & 0.75 \\
			15   & 0.70 & 0.66 & 0.72 & \textbf{0.74} & 0.73 \\
			13.5 & 0.66 & 0.58 & 0.66 & 0.68 & \textbf{0.70} \\
			12.5 & 0.62 & 0.53 & 0.61 & 0.66 & \textbf{0.67} \\
			12   & 0.60 & 0.49 & 0.59 & 0.64 & \textbf{0.66} \\
			11   & 0.56 & 0.43 & 0.53 & 0.58 & \textbf{0.62} \\
			10.5 & 0.53 & 0.40 & 0.51 & 0.56 & \textbf{0.60} \\
			\hline
		\end{tabular}
	\end{center}
	\label{tab_salinas}
\end{table}	
Denoising results for $Subject_{1}$ image from the IRIS Lab multispectral face image are illustrated in Fig. \ref{subject1} with input PSNR=19 dB. Performance evaluation  is shown in Fig. \ref{curve_subject1}. It demonstrates the improvement of OVNLM over the other algorithms with different noise levels and with distinctive performance in low PSNR (11 dB and 10.5 dB). SSIM results presented in Tab. \ref{tab_sub1} also demonstrate the outperformance of OVNLM. With low PSNR, M-SURE exhibits the lowest performance with SSIM=0.47 while with OVNLM we have SSIM=0.63.
Fig. \ref{subject2} shows the  outputs of the denoising algorithms with input PSNR=19 dB for $Subject_2$. OVNLM presents the best performance with 33.07 dB compared to 28.4 dB for M-SURE, 27.7 dB for MNLM, 27.5 dB for BM4D and 25 dB for TDL. This performance is confirmed by Fig. \ref{curve_subject2} which illustrates the PSNR variations with respect to the input PSNR. SSIM results in Tab. \ref{tab_sub2} confirm the outperformance of OVNLM particularly in presence of high noise level. For example, with input PSNR=10.5 dB, we have SSIM=0.67 for OVNLM which is better than the rest of the algorithms.
\begin{table}[!h]
	\caption{SSIM results for $Subject_{1}$. The best results are shown in boldface. with relatively high PSNR, M-SURE presents the best performance. With high noise level, OVNLM exhibits the best results. Its SSIM values are far better than the other algorithms.}
	\begin{center}
		\begin{tabular}{ c c c c c c c}
			\hline
			Input \newline PSNR & MNLM & M-SURE & BM4D & TDL & OVNLM \\
			\hline
			22.5 & 0.88 & \textbf{0.90} & 0.87 & 0.85 & 0.86 \\
			19   & 0.82 & \textbf{0.85} & 0.82 & 0.79 & 0.82 \\
			17   & 0.75 & \textbf{0.80} & 0.77 & 0.74 & 0.78 \\
			15   & 0.75 & 0.73 & 0.73 & 0.68 & \textbf{0.76} \\
			13.5 & 0.69 & 0.66 & 0.69 & 0.65 & \textbf{0.73} \\
			12.5 & 0.64 & 0.61 & 0.66 & 0.61 & \textbf{0.70} \\
			12   & 0.62 & 0.58 & 0.64 & 0.58 & \textbf{0.69} \\
			11   & 0.56 & 0.51 & 0.60 & 0.55 & \textbf{0.65} \\
			10.5 & 0.53 & 0.47 & 0.58 & 0.51 & \textbf{0.63} \\
			\hline
		\end{tabular}
	\end{center}
	\label{tab_sub1}
\end{table}
\begin{table}[!h]
	\caption{SSIM results for $Subject_{2}$. The best results are shown in boldface: with relative high PSNR, M-SURE presents the best performance. With high noise levels, OVNLM exhibits the best results. Its SSIM values are far better than the other algorithms.}
	\begin{center}
		\begin{tabular}{  c c c c c c   }
			\hline
			Input  PSNR & MNLM & M-SURE & BM4D & TDL & OVNLM \\
			\hline
			22.5 & 0.86 & \textbf{0.88} & 0.87 & 0.87 & \textbf{0.88} \\
			19   & 0.82 & 0.82 & 0.82 & 0.83 & \textbf{0.84} \\
			17   & 0.78 & 0.77 & 0.80 & 0.77 & \textbf{0.81} \\
			15   & 0.72 & 0.73 & 0.76 & 0.72 & \textbf{0.79} \\
			13.5 & 0.67 & 0.66 & 0.68 & 0.72 & \textbf{0.76} \\
			12.5 & 0.63 & 0.61 & 0.62 & 0.70 & \textbf{0.74} \\
			12   & 0.60 & 0.58 & 0.61 & 0.68 & \textbf{0.72} \\
			11   & 0.56 & 0.51 & 0.57 & 0.62 & \textbf{0.69} \\
			10.5 & 0.53 & 0.47 & 0.54 & 0.60 & \textbf{0.67} \\
			\hline
		\end{tabular}
	\end{center}
	\label{tab_sub2}
\end{table}
\newline
Fig. \ref{sub1_8} shows PSNR variations of $Subject_{i}$  $(i=3 \ldots 8)$ from IRIS Lab multispectral face image database. These results confirm the outperformance of OVNLM algorithm over the other algorithms. This performance is emphasized by the SSIM results for $Subject_{i}$  $(i=3 \ldots 8)$ presented in Tab. \ref{tab_sub3,tab_sub4,tab_sub5,tab_sub6,tab_sub7,tab_sub8}. According to these SSIM values, denoised images obtained with OVNLM are the most similar ones to the original images. \newline
\noindent To emphasize the importance of a good parametrization of the proposed filter, we analyzed its performance by varying parameter $h$ which controls the degree of smoothing for a fixed covariance matrix $\Phi$  with different noise levels as shown in Fig. \ref{curve_h} for $Subject_9$ multispectral image. We noticed that the resulting PSNR depends enormously on the choice of parameter $h$ which also implies a good choice of parameter $\Phi$. 
\newline
The overall experimental results confirm the following findings. First, NLM based algorithm (MNLM) without a proper choice of parameters cannot deal with high level noise. Second, dictionary methods are known  by their lack of robustness which is confirmed by TDL results. M-SURE, a wavelet-based method and BM4D, a grouping and collaborative filtering method, do not exhibit stable denoising performance. Furthermore, SSIM results of M-SURE showed that this algorithm does not preserve details. Indeed, for $Subject_4$ and with PSNR=10.5, M-SURE exhibited the lowest value which is too much smaller than BM4D, TDL and OVNLM values. For BM4D, some details such as the ones of Salinas image in Fig. \ref{salina_results} are lost. The experimental findings reflect also the intrinsic characteristics of the proposed method: (i) OVNLM performs better in term of details preservation. This is quantitatively confirmed by the SSIM results. (ii) The fine tuning of the parameters allows OVNLM to achieve better denoising performance particularly with high level noise. Indeed, by comparing OVNLM results to MNLM, we deduce the  importance of setting up a proper choice for these parameters. (iii) Our relevant pixel selection strategy conserves the non-local property but chooses the most influential pixels for each pixel restoration. PSNR results have demonstrated the effectiveness of our approach. Next, we focus on the computation complexity of our method and compare it with other NLM based algorithms.
	\subsection{Computation Complexity}
	To evaluate the relevance of choosing a particular subset for each pixel in term of computation complexity, we evaluate the computation time of the OVNLM algorithm and compare it against the computation time of several variations of the NLM filter: classic NLM filter applied on each spectral band separately, the MNLM algorithm, the proposed OVNLM algorithm applied with and without selecting subsets for each pixel and referred as OVNLM v1 and OVNLM v2 respectively.
	We use the multispectral image of $Subject_{1}$  with input PSNR=22.5 dB. We test the algorithms on a Windows machine, Intel Core i7, 3.1 GHz with 8GB RAM. Results are illustrated in Table \ref{tab_time}. \newline
	\begin{table}[!b]
		\caption{Computational time and denoising quality for different denoising algorithms applied on multispectral image of $Subject_{1}$ }
		\centering
		\begin{tabular}{c c c}
			\hline
			Method & Computational time (s) & PSNR \\
			\hline
			NLM      & 195  & 30.21 \\
			MNLM     & 35   & 30.29 \\
			OVNLM v1 & 29   & 32.03 \\
			OVNLM v2 & 175  & 33.1 \\
			\hline
		\end{tabular}
		\label{tab_time}
	\end{table}
\noindent First of all, we notice that MNLM and OVNLM v1 exhibit the best performance in term of computational time. Classic NLM filter and OVNLM v2 need to consider the contribution of every pixel in the image domain which justifies their high computational time. Results also demonstrate that the computational reduction that characterizes OVLM v2 comes with a slight decrease of PSNR compared to OVNLM v2. SURE procedure itself does not greatly affect the computation time which is also confirmed in \citep{6648714,5165022}.\newline
We analyze the effect of the choice of the parameter $\varsigma$ which controls the amount of points to be considered when restoring each pixel. We experiment on $Subject_{10}$ multispectral image. The findings in Tab. \ref{tab_param} show that at a certain level ($\varsigma  \ge $ 100), we achieve a considerable improvement in term of computational time. Meanwhile, the improvement in term of PSNR is not very relevant (around 1dB). We conclude  that a proper adjustment of this parameter can ensure a good denoising performance while keeping a reasonable computational time. 
	\begin{table}[!b]
		\caption{Variation of PSNR and computational time with respect to parameter $\varsigma$  }
		\centering
		\begin{tabular}{c c c c c c c c}
			\hline
			$\varsigma$ & 1 & 10 & 50 & 100 & 500 & $10^3$ & $10^4$ \\
			\hline
			PSNR      & 27.38  & 28.52 & 29.76& 32.88& 33.19&33.34 & 33.83\\
			Time (s)     & 20   & 33 & 43 & 47 &57 &60 &68 \\

			\hline
		\end{tabular}
		\label{tab_param}
	\end{table}

Finally, we can state that the proposed denoising filter demonstrated its effectiveness  compared to other algorithms. Experiments on real multispectral images have shown good results in terms of PSNR. While denoising, we tried to take advantage of the spectral information by extending the NLM filter to the vector case, optimized the choice of its parameter and reduced the computation complexity.
	\section{Conclusion}
	In this paper, we have proposed a novel multispectral image denoising algorithm. The proposed scheme is an improvement of Non-Local Mean filter. This improvement is obtained by extending NLM filter to the vector case. \newline
	However, the choice of the filter tuning parameters still poses a challenge. In this work,
	filter parameters are tuned automatically using an optimization technique based on SURE. 
	Indeed, unbiased risk estimator is applied and filter parameters are obtained by minimizing the expression of SURE. An easily sequential quadratic programming is used to solve the non-linear minimization problem.\newline
	Experiments performed on color and multispectral face images demonstrate the superiority of the proposed framework compared with two other well-known similar algorithms. 
	Good performance in terms of PSNR and SSIM is obtained. Nevertheless, the computation burden is still an important challenge with NLM in general. Further techniques to speed up the computation should be investigated.
	\section*{Acknowledgment}
	This publication was made possible by NPRP grant \# 4-1165-
	2-453 from the Qatar National Research Fund (a member of
	Qatar Foundation). The statements made herein are solely the
	responsibility of the authors.
		\begin{table}[!h]
			\caption{SSIM results for $Subject_{3}$. The best results are bold.}
			\begin{center}
				\begin{tabular}{   c  c  c  c  c  c   }
					\hline
					Input \newline PSNR & MNLM & M-SURE & BM4D & TDL & OVNLM \\ \hline
					22.5 & 0.86 & 0.87 & \textbf{0.88} & 0.85 & \textbf{0.88} \\
					19  & 0.81 & 0.79 & 0.83 & 0.80 & \textbf{0.84} \\
					17  & 0.76 & 0.73 & 0.80 & 0.76 & \textbf{0.82} \\
					15  & 0.70 & 0.66 & 0.77 & 0.72 & \textbf{0.79} \\
					13.5 & 0.64 & 0.60 & 0.72 & 0.64 & \textbf{0.77} \\
					12.5 & 0.59 & 0.54 & 0.69 & 0.63 & \textbf{0.74} \\
					12  & 0.56 & 0.51 & 0.68 & 0.59 & \textbf{0.72} \\
					11  & 0.52 & 0.45 & 0.63 & 0.58 & \textbf{0.69} \\
					10.5 & 0.49 & 0.41 & 0.60 & 0.55 & \textbf{0.67} \\ \hline
				\end{tabular}
			\end{center}
			\label{tab_sub3}
		\end{table}
			\begin{table}[!h]
				\caption{SSIM results for $Subject_{4}$. The best results are bold.}
				\begin{center}
					\begin{tabular}{   c  c  c  c  c  c   }
						\hline
						Input PSNR & MNLM & M-SURE & BM4D & TDL & OVNLM \\ \hline
						22.5 & 0.85 & 0.86 & \textbf{0.87} & 0.85 & \textbf{0.87} \\
						19  & 0.80 & 0.79 & \textbf{0.83} & 0.79 & \textbf{0.83} \\
						17  & 0.75 & 0.65 & 0.79 & 0.76 & \textbf{0.81} \\
						15  & 0.68 & 0.57 & 0.75 & 0.69 & \textbf{0.77} \\
						13.5 & 0.62 & 0.56 & 0.70 & 0.64 & \textbf{0.73} \\
						12.5 & 0.57 & 0.50 & 0.67 & 0.62 & \textbf{0.69} \\
						12  & 0.55 & 0.47 & 0.65 & 0.58 & \textbf{0.67} \\
						11  & 0.50 & 0.40 & 0.60 & 0.57 & \textbf{0.61} \\
						10.5 & 0.47 & 0.37 & 0.57 & 0.56 & \textbf{0.59} \\ \hline
					\end{tabular}
				\end{center}
				\label{tab_sub4}
			\end{table}
			\begin{table}[!h]
				\caption{SSIM results for $Subject_{5}$. The best results are bold.}
				\begin{center}
					\begin{tabular}{  c  c  c  c  c  c   }
						\hline
						Input PSNR & MNLM & M-SURE & BM4D & TDL & OVNLM \\ \hline
						22.5 & 0.89 & 0.89 & 0.89 & 0.88 & \textbf{0.90}\\ 
						19  & 0.85 & 0.84 & 0.86 & 0.83 & \textbf{0.88}\\ 
						17  & 0.81 & 0.79 & 0.82 & 0.79 & \textbf{0.86}\\ 
						15  & 0.76 & 0.72 & 0.78 & 0.75 & \textbf{0.81}\\ 
						13.5 & 0.70 & 0.66 & 0.75 & 0.70 & \textbf{0.80}\\ 
						12.5 & 0.66 & 0.61 & 0.72 & 0.65 & \textbf{0.78}\\ 
						12  & 0.64 & 0.58 & 0.69 & 0.64 & \textbf{0.76}\\ 
						11  & 0.58 & 0.51 & 0.65 & 0.61 & \textbf{0.71}\\ 
						10.5 & 0.56 & 0.48 & 0.64 & 0.60 & \textbf{0.69}\\ \hline
					\end{tabular}
				\end{center}
				\label{tab_sub5}
			\end{table}
			\begin{table}[!h]
				\caption{SSIM results for $Subject_{6}$. The best results are bold.}
				\begin{center}
					\begin{tabular}{   c  c  c  c  c  c   }
						\hline
						Input PSNR & MNLM & M-SURE & BM4D & TDL & OVNLM \\ \hline
						22.5 & 0.87 & \textbf{0.89} & \textbf{0.89} & 0.88 & \textbf{0.89}\\ 
						19  & 0.83 & 0.83 & 0.85 & 0.83 & \textbf{0.86}\\
						17  & 0.79 & 0.78 & 0.82 & 0.79 & \textbf{0.83}\\ 
						15  & 0.74 & 0.72 & 0.77 & 0.74 & \textbf{0.80}\\ 
						13.5 & 0.69 & 0.66 & 0.74 & 0.70 & \textbf{0.77}\\
						12.5 & 0.65 & 0.61 & 0.72 & 0.66 & \textbf{0.74}\\ 
						12  & 0.63 & 0.59 & 0.69 & 0.62 & \textbf{0.72}\\ 
						11  & 0.58 & 0.53 & 0.65 & 0.58 & \textbf{0.68}\\ 
						10.5 & 0.56 & 0.50 & 0.62 & 0.57 & \textbf{0.66}\\ \hline
					\end{tabular}
				\end{center}
				\label{tab_sub6}
			\end{table}	
			\begin{table}[!h]
				\caption{SSIM results for $Subject_{7}$. The best results are bold.}
				\begin{center}
					\begin{tabular}{   c  c  c  c  c c   }
						\hline
						Input PSNR & MNLM & M-SURE & BM4D & TDL & OVNLM \\ \hline
						22.5 & 0.89 & \textbf{0.90} & 0.89 & 0.89 & \textbf{0.90}\\
						19  & 0.85 & 0.84 & 0.85 & 0.83 & \textbf{0.87}\\ 
						17  & 0.82 & 0.80 & 0.83 & 0.78 & \textbf{0.85}\\ 
						15  & 0.77 & 0.74 & 0.78 & 0.75 & \textbf{0.82}\\ 
						13.5 & 0.72 & 0.66 & 0.74 & 0.70 & \textbf{0.78}\\ 
						12.5 & 0.68 & 0.64 & 0.72 & 0.67 & \textbf{0.77}\\ 
						12  & 0.66 & 0.62 & 0.69 & 0.64 & \textbf{0.75}\\ 
						11  & 0.61 & 0.56 & 0.64 & 0.58 & \textbf{0.71}\\ 
						10.5 & 0.58 & 0.52 & 0.63 & 0.58 & \textbf{0.68}\\ \hline
					\end{tabular}
				\end{center}
				\label{tab_sub7}
			\end{table}		
			\begin{table}[!h]
				\caption{SSIM results for $Subject_{8}$. The best results are bold.}
				\begin{center}
					\begin{tabular}{   c  c  c  c  c  c   }
						\hline
						Input PSNR & MNLM & M-SURE & BM4D & TDL & OVNLM \\ \hline
						22.5 & 0.88 & \textbf{0.89} & \textbf{0.89} & 0.88 & 0.88\\ 
						19  & 0.83 & 0.83 & \textbf{0.85} & 0.83 & \textbf{0.83}\\ 
						17  & 0.79 & 0.78 & \textbf{0.82} & 0.79 & \textbf{0.82}\\ 
						15  & 0.72 & 0.70 & 0.78 & 0.73 & \textbf{0.79}\\ 
						13.5 & 0.66 & 0.63 & 0.74 & 0.68 & \textbf{0.75}\\ 
						12.5 & 0.61 & 0.57 & 0.71 & 0.66 & \textbf{0.73}\\ 
						12  & 0.59 & 0.54 & 0.70 & 0.62 & \textbf{0.71}\\ 
						11  & 0.53 & 0.47 & 0.63 & 0.58 & \textbf{0.67}\\ 
						10.5 & 0.50 & 0.43 & 0.62 & 0.55 & \textbf{0.64}\\ \hline
					\end{tabular}
				\end{center}
				\label{tab_sub8}
			\end{table}	
			
	\clearpage
	
	\section*{Appendix}
	Let: $\chi(p)=exp\left(-\frac{1}{h^2}\sum_{k\in K} (y_{s-k}-y_{p-k})^{T}\Phi^{-1}(y_{s-k}-y_{p-k}) \right)$.
	\begin{equation*}
	\begin {tabular}{l}
	\(\bullet p-s \in K: \)
	\\
	\\
	\( =exp(-\frac{1}{h^2}(y_s - y_p)^T\Phi^{-1}(y_s - y_p)) \cdot exp(-\frac{1}{h^2}(y_{2s-p} - y_s)^T\Phi^{-1}(y_{2s-p} - y_s)) \cdot  \)\\
	\\
	\(exp(-\frac{1}{h^2} \sum_{\substack{k\in K\\ k\not\in \{0,p-s\}}} (y_{s-k}-y_{p-k})^{T}\Phi^{-1}(y_{s-k}-y_{p-k}))     \)\\
	\\
	\(\Longrightarrow  \frac{\partial \chi(p)}{\partial y_{s_j}}=exp(-\frac{1}{h^2} \sum_{\substack{k\in K\\ k\not\in \{0,p-s\}}} (y_{s-k}-y_{p-k})^{T}\Phi^{-1}(y_{s-k}-y_{p-k})) \cdot\)
	\\
	\\
	\(\frac{\partial}{\partial y_{s_j}} \left( exp(-\frac{1}{h^2}(y_s - y_p)^T\Phi^{-1}(y_s - y_p))  \cdot exp(-\frac{1}{h^2}(y_{2s-p} - y_s)^T\Phi^{-1}(y_{2s-p} - y_s)) \right) \)\\
	\\
	\( \frac{\partial \chi(p)}{\partial y_{s_j}}=\chi(p) \left[ \left( (y_p -y_s)^T\frac{1}{h^2}(\Phi^{-1} + {\Phi^{-1}}^T) \right)^T_j + \left( (y_s -y_{2s-p})^T\frac{1}{h^2}(\Phi^{-1} + {\Phi^{-1}}^T) \right)^T_j \right]\)
	\\
	\\
	\(\bullet p-s \not\in K: \)
	\\
	\\
	\( =exp(-\frac{1}{h^2}(y_s - y_p)^T\Phi^{-1}(y_s - y_p)) \cdot  \)\\
	\\
	\(exp(-\frac{1}{h^2} \sum_{\substack{k\in K\\ k\neq 0}} (y_{s-k}-y_{p-k})^{T}\Phi^{-1}(y_{s-k}-y_{p-k}))     \)\\
	\end{tabular}
	\end{equation*}
	\begin{equation*}
	\begin {tabular}{l}
	\(\Longrightarrow  \frac{\partial \chi(p)}{\partial y_{s_j}}=exp(-\frac{1}{h^2} \sum_{\substack{k\in K\\ k\neq 0}} (y_{s-k}-y_{p-k})^{T}\Phi^{-1}(y_{s-k}-y_{p-k})) \cdot\)
	\\
	\\
	\(\frac{\partial}{\partial y_{s_j}} \left( exp(-\frac{1}{h^2}(y_s - y_p)^T\Phi^{-1}(y_s - y_p))  \right) \)\\
	\\
	\( \frac{\partial \chi(p)}{\partial y_{s_j}}=\chi(p)  \left( (y_p -y_s)^T\frac{1}{h^2}(\Phi^{-1} + {\Phi^{-1}}^T) \right)^T_j  \)
	
	\end{tabular}
	\end{equation*}
	\section*{References}
\bibliographystyle{model2-names}

\end{document}